\documentclass[10pt]{article}

\usepackage[utf8]{inputenc}
\usepackage{times}
\usepackage[top=1in, bottom=1in, left=1.5in, right=1.5in]{geometry}

\usepackage{multicol}
\usepackage{graphicx}
\graphicspath{{academic/figs/}}

\usepackage{dblfloatfix}
\usepackage[ruled]{algorithm2e}

\SetCommentSty{mycommfont}
\SetFuncSty{myprocname}

\usepackage{changepage}

\usepackage{mathtools}
\usepackage{mathrsfs}
\usepackage{amsmath, amsthm}
\usepackage{bm}
\usepackage{bbm}
\usepackage{amssymb}

\usepackage[numbers]{natbib}
\usepackage{blindtext}

\usepackage{upgreek}
\usepackage{multirow}
\usepackage{amsfonts}
\usepackage{here}
\usepackage[normalem]{ulem}
\usepackage{booktabs}
\usepackage{color}
\usepackage{subcaption}
\usepackage{minitoc}
\usepackage[font=small, skip=5pt]{caption}
\usepackage[shortlabels]{enumitem}
\usepackage{listings}
\usepackage{xurl}
\usepackage{tikz}
\usepackage[bookmarks=true]{hyperref}
\usepackage{csquotes}
\usepackage{xcolor} 
\hypersetup{
    colorlinks,
    linkcolor={red!50!black},
    citecolor={blue!80!black},
    urlcolor={blue!80!black}
}

\usepackage{cleveref}
\usepackage{pdfpages}

\renewenvironment{abstract}{%
\vspace{-0.45in}\hfill
\begin{center}
\begin{minipage}{0.75\textwidth}
\textbf{Abstract:}
}
{
\end{minipage}
\end{center}
}

\newcommand{\E}{\mathbb{E}}

\newcommand{\V}{\mathsf{V}}

\renewcommand{\d}{\partial}
\newcommand{\D}{\Delta}
\newcommand{\e}{\text{e}}

\definecolor{purple}{rgb}{0.858, 0.08, 0.85}
\definecolor{darkgreen}{rgb}{0.08, 0.55, 0.08}
\definecolor{blue1}{rgb}{0.2, 0.2, 0.6}
\definecolor{green1}{rgb}{0.2, 0.6, 0.2}
\definecolor{green2}{rgb}{0.1, 0.4, 0.1}
\definecolor{dkgreen}{rgb}{0,0.6,0}
\definecolor{gray}{rgb}{0.5,0.5,0.5}
\definecolor{mauve}{rgb}{0.58,0,0.82}

\makeatletter
\newcommand*\bigcdot{\mathpalette\bigcdot@{.5}}
\newcommand*\bigcdot@[2]{\mathbin{\vcenter{\hbox{\scalebox{#2}{$\m@th#1\bullet$}}}}}
\makeatother

\title{\textbf{Sample Efficient Robot Learning with Structured World Models}}
\author{
  Tuluhan Akbulut *
  \and
  Max Merlin *
    \and
  Shane Parr *
    \and
  Benedict Quartey *
    \and
  Skye Thompson
  \footnote{All authors contributed equally to this work.}
}
\date{December 16, 2021}

\begin{document}

\maketitle
\begin{abstract}

Reinforcement learning has been demonstrated as a flexible and effective approach for learning a range of continuous control tasks, such as those used by robots to manipulate objects in their environment. But in robotics particularly, real-world rollouts are costly, and sample efficiency can be a major limiting factor when learning a new skill. In game environments, the use of world models has been shown to improve sample efficiency while still achieving good performance, especially when images or other rich observations are provided. In this project, we explore the use of a world model in a deformable robotic manipulation task, evaluating its effect on sample efficiency when learning to fold a cloth in simulation. We compare the use of RGB image observation with a feature space leveraging built-in structure (keypoints representing the cloth configuration), a common approach in robot skill learning, and compare the impact on task performance and learning efficiency with and without the world model. Our experiments showed that usage of keypoints increased  performance of the best model on the task by 50\%, and in general, the use of a learned or constructed reduced feature space improved task performance and sample efficiency. The use of a state transition predictor(MDN-RNN) in our world models did not have a notable effect on task performance.
\end{abstract}

\section*{Introduction}

Reinforcement Learning (RL) is a policy-learning approach that incentivizes an agent to produce desired behavior through feedback (reward) from the environment conditioned on actions taken by said agent\cite{sutton2018reinforcement}.This formulation has appealing flexibility and ability to generalize to a wide class of problems, from structured games to continuous control. However, one limitation of RL algorithms in complex environments, where the state and action space is large and reward may be sparse, is their sample inefficiency. Capturing sufficient information about an environment to learn a useful policy can require taking hundreds of thousands of actions and recording the results. This is particularly problematic when ‘samples’ are hard to come by - for instance, on problems in real-world environments, like robotics, where taking such actions are costly \cite{dulacarnold2019challenges}. 

 World Models (WMs) attempt to improve the sample efficiency of RL approaches by learning a model that captures the temporal and spatial dynamics of the environment, \cite{Jurgen} so that an agent can learn a policy by leveraging that model to reduce the number of interactions it needs to take in the real world. A world model learns the dynamics of an environment independent of a specific policy the agent might take, allowing it to leverage information gained through off-policy exploration, and to simulate the effect of taking actions in states that may be otherwise hard to reach. Some recent RL approaches using world models \cite{Dreamerv2} have been shown to improve sample efficiency over state-of-the-art  in atari benchmarks. In this project, we investigate the performance of one world model approach on a continuous control task in a complex environment with a large state space: a robot manipulating a cloth to fold it in simulation. 
 
 We found that the learning state representations improved the learning performance but the environment dynamics predictor of world models did not improve the performance over our ablation experiments. However, we should note that the approach we used is not the current state-of-the-art, and our experimentation was limited by computing constraints. Secondly, we show that using structured data as states, i. e. keypoints instead of RGB images, increases both performance and sample efficiency more than learned state representations. This suggests that it is possible to increase sample efficiency of world models with the use of a structured feature space - but more progress is needed for world models to be effectively used to solve real-world robotics tasks.

\section*{Related Work}

Ha and Schmidhuber \cite{Jurgen} demonstrate the effectiveness of a learned generative world model on a simulated racecar task and Doom, learning to encode a pixel image of the state as a latent representation from which the next state can be reconstructed, and learning a separate policy model to select the best action given the encoded state. \citet{hafner2020dream} tries WMs in DeepMind Control Suite \cite{tassa2018deepmind} with a more complex policy model. \citet{Dreamerv2} discretize the latent space of WMs to increase their prediction performance and their model achieves a better performance than state-of-the-art RL models and humans in atari benchmark. In \citet{kipf2020contrastive}, structured world models are learned, although they are not used for exploration. Finally, world models have been used for exploration via planning by Sekar et al. \cite{sekar2020planning}. Our project expands on the literature by introducing built-in structure to the state space and better exploration strategies to make WMs a better fit for robotics. Our approach is directly informed by that of Ha and Schmidhuber, though we make adjustments to our model training approach based on what has been shown to be effective in later literature.\cite{Dreamerv2}

\section*{Problem Statement}

We assume a task in an environment that can be modeled as an MDP, with a state $s$ from a state space $S$, where the agent selects an action $a$ from its action space $A$ at each timestep, transitioning to a state $s'$ and receiving a reward $r$. The agent's goal is to select actions that maximize the cumulative discounted reward given the agent's starting state. Given a dataset collected in the environment with some exploration policy, consisting of state transitions, actions taken, and rewards, we aim to learn a model capturing that MDP's transition function: $s_t, a_t \xrightarrow[]{} s_{t+1}$. At test time, the agent uses this model to select $a_t$ with the highest predicted Q-value. The task we attempt to learn, cloth folding, is a continuous control task with a large and complex state space, due to the infinitely many possible configurations of the cloth. Learning a policy directly in this environment, without prior structure, knowledge of environment dynamics, or bias towards meaningful state features or regions of the state space, would be very sample intensive.

Instead of mapping the observed state of the real world directly to an action, we want to find a compact state representation that captures temporal and spatial environment dynamics. The idea is that this reduced representation includes relevant and useful information for the task, and allows for more efficient learning due to its smaller space of relevant features. These representations are used to learn a transition model that captures environment dynamics. A controller is then trained to select optimal actions using this state representation and knowledge of possible state transitions. 

\section*{World Models for Efficient Robot Learning}

Our base model consists of three components, learning a reduced state representation, dynamics model, and controller respectively.

\begin{figure}[t]
\includegraphics[width=0.95\textwidth]{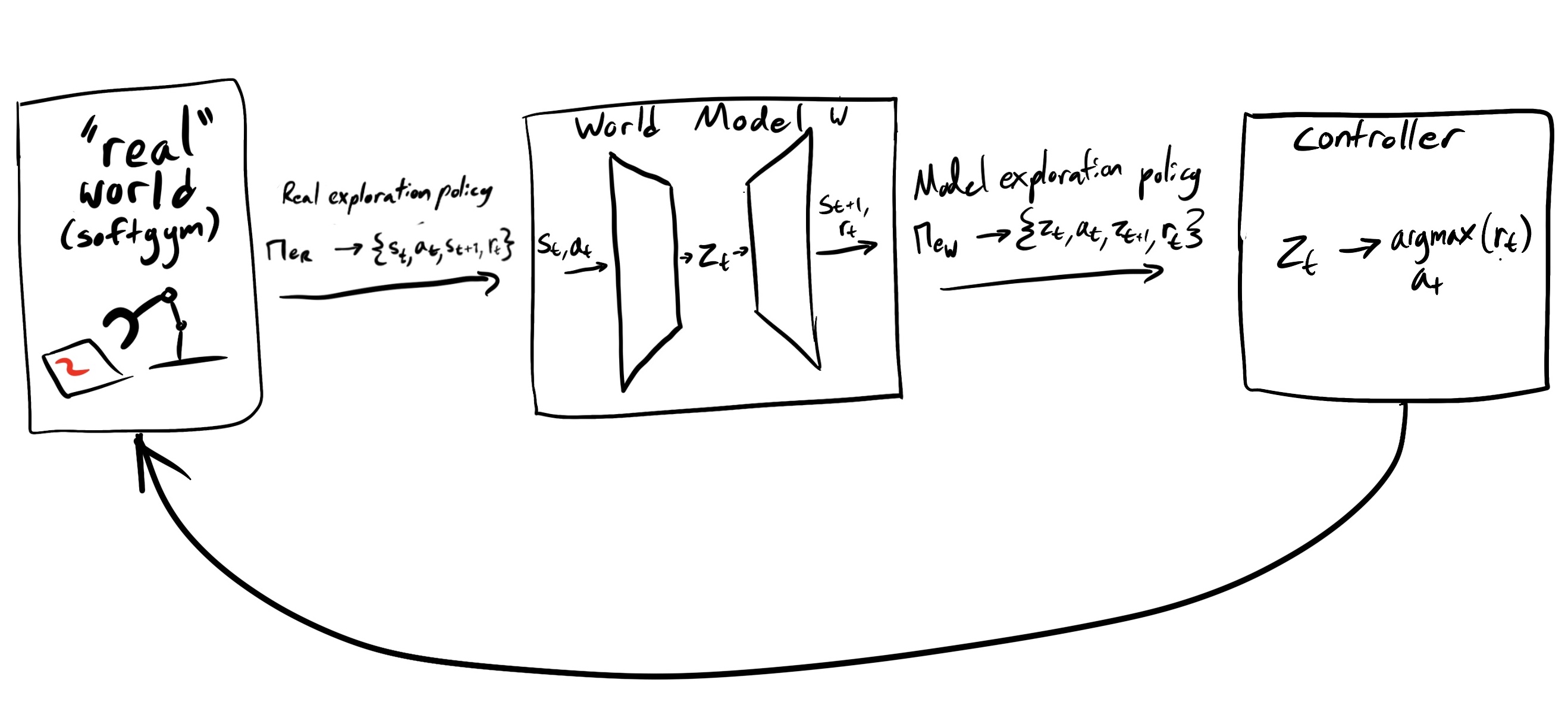}
\caption{Base world model. An autoencoder is used to learn a reduced latent representation of the state. An RNN is then trained to predict the latent representation of the state at the next timestep, outputting a hidden representation vector. This information is used to train a controller. The whole system is trained iteratively on the rollouts produced by controller training, for more efficient exploration and learning.}
\end{figure}

\subsection*{State Representation}

The simulator environment we use in our experiments provides a top-down view of the world captured by a fixed camera, in the form of 3x64x64 RGB images. This representation provides nearly all the meaningful state information for our task at each timestep, but is difficult to exploit efficiently due to its high dimensionality. We investigate two approaches for extracting a reduced state representation - learning a latent feature space by training a Variational Autoencoder(VAE) \cite{Diederik2019} on images of the environment, and extracting a lower-dimensional structured representation of the state in the form of keypoints. These keypoints are 24 points uniformly distributed in a grid over the surface of the cloth, which we calculate using a mesh representation of the cloth accessible in the simulation environment.

Using these keypoint positions as a state representation exploits known information about task structure, meaning they carry necessary and sufficient information for cloth folding, whereas inferring a similar knowledge from the scene image is an additional hurdle. This is a common concession to the difficulty of learning continuous control in robotics, and much work has been done on what object representations can be leveraged for more effective and efficient skill learning, and how those representations can be efficiently constructed \cite{structure}. We used the ground truth keypoint positions for this project - on a real robot, we would use some pretrained model capable of estimating keypoint positions from depth images such as \citet{florence2018dense} or \citet{gao2021kpam}. This would be a noisier state signal than our observations in simulation, but has been an effective enough approximation for successful learning in other experiments \cite{kulkarni2019unsupervised}.

\subsection*{Dynamics Model}

Given some representation of the state of the world and an action taken, we then train a recurrent neural network (RNN) to attempt to predict the state representation at the next timestep, along with outputting a hidden state that captures any non-markovian properties representing world dynamics (velocity, acceleration, etc) that may be relevant to that transition. This allows us to "imagine" the effect of an action in the world, capturing information to better estimate the action that would maximize cumulative reward, and enabling the agent to behave more optimally.

For our model, we use a mixture density recurrent neural network (MDN-RNN), which predicts a Gaussian mixture model over the possible future states, from which a representation of the predicted next state may be sampled. This allows the model to represent stochastic environments, as well as the ability to adjust confidence in the model's representation through the use of a temperature parameter $\tau$, which when increased, constrains the distribution on Gaussians in the mixture model to be more uniform, leading to a higher variance in the predicted next state. For all our experiments, given that our task environment is deterministic, we use $\tau = 1.0$.

\subsection*{Controller}

Our controller model is a linear model that predicts the action that will maximize future cumulative reward given the current state representation $z_t$ and hidden state representation $h_t$ output by the RNN. It consists of only 867 parameters– the simplicity of this model helps validate the effects our state representation and dynamics model has on task completion. The small parameter size allows us to train a policy using an Evolutionary Strategy (ES) instead of gradient descent. Our chosen algorithm is called Covariance Matrix Adaptation Evolution Strategy (CMA-ES) \cite{hansen2016cma}, where a population of solutions are sampled from a multivariate Gaussian distribution, and the best $x\%$ of the solutions are used to define the new distribution for the next population generation, where $x$ is an hyperparameter. Differently from other ES algorithms like Cross Entropy Method (CEM) \cite{Boer04atutorial}, CMA-ES successfully adapts the covariance matrix by weighting solutions differently according to performance, and the covariance matrix is calculated with these weights for adaptation. In addition, the mean of the old distribution before elimination of bad solutions is used for covariance matrix calculation for better adaptation.

\subsection*{Training procedure}

One challenge when training a world model in a complex, high-dimensional environment is encountering relevant and important states, which may be sparse or difficult to reach, in order to properly capture their dynamics. Simply training on random rollouts in the world may not be enough to achieve this efficiently. We mitigate this problem by training our models iteratively. In our first pass through the world model and controller model, we train on a set of 1500 rollouts of a controller that selects actions at random. We sequentially train the VAE and RNN models on these, then run 10 generations of the controller and record each rollout generated, producing ~200 more samples each training iteration. These samples, combined with 50 rollouts randomly selected from all previously generated data, are used to train and validate all models in the subsequent training iteration. This biases the learned representations towards the improved rollouts that interact more with the cloth in later training iterations as the entire system trains.

\section*{Software and Environment}

\begin{figure}[H]
 \begin{minipage}{.5\textwidth}
 \centering
\includegraphics[width=.6\textwidth]{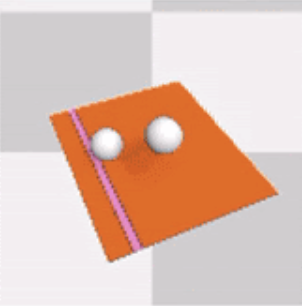}
\caption{The initial frame from the fold cloth task.}
\label{fig:initial_sim}
\end{minipage}
 \begin{minipage}{.5\textwidth}
 \centering
\includegraphics[width=.6\textwidth]{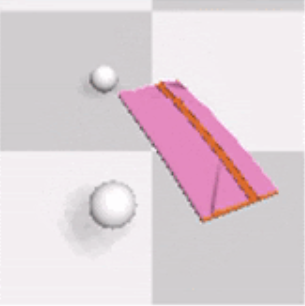}
\caption{A frame of the completed task.}
\label{fig:end_sim}
\end{minipage}
\end{figure}

We used softgym \cite{softgym}, a simulator environment designed for learning deformable manipulation skills. We selected a cloth folding task because it is a complicated task with a meaningful keypoint representation, and the cloth is easier for the VAE to detect and reconstruct due its relatively large size and distinct color.The environment provided the RGB images and position of two grippers in 3D space (represented by the white spheres), and a binary variable saying whether the gripper was picking up the cloth at that timestep.
The cloth always starts in the same initial state, centered in the view of the RGB camera (Figure \ref{fig:initial_sim}). The reward for the task at each time step was the negative summed distance between a grid of points on each half of the cloth, along with an additional negative penalty of how far the cloth was from the center of the camera view (to keep the grippers from dragging the cloth off screen when folding it. The action space was the displacement vector at a timestep provided to the gripper, indicating the magnitude and direction it should move at the next timestep. Figure \ref{fig:end_sim} illustrates the end frame of a successful task execution.

\section*{Experiments}

\begin{figure}[H]
\centering
\begin{minipage}{.9\textwidth}
\includegraphics[width=\textwidth]{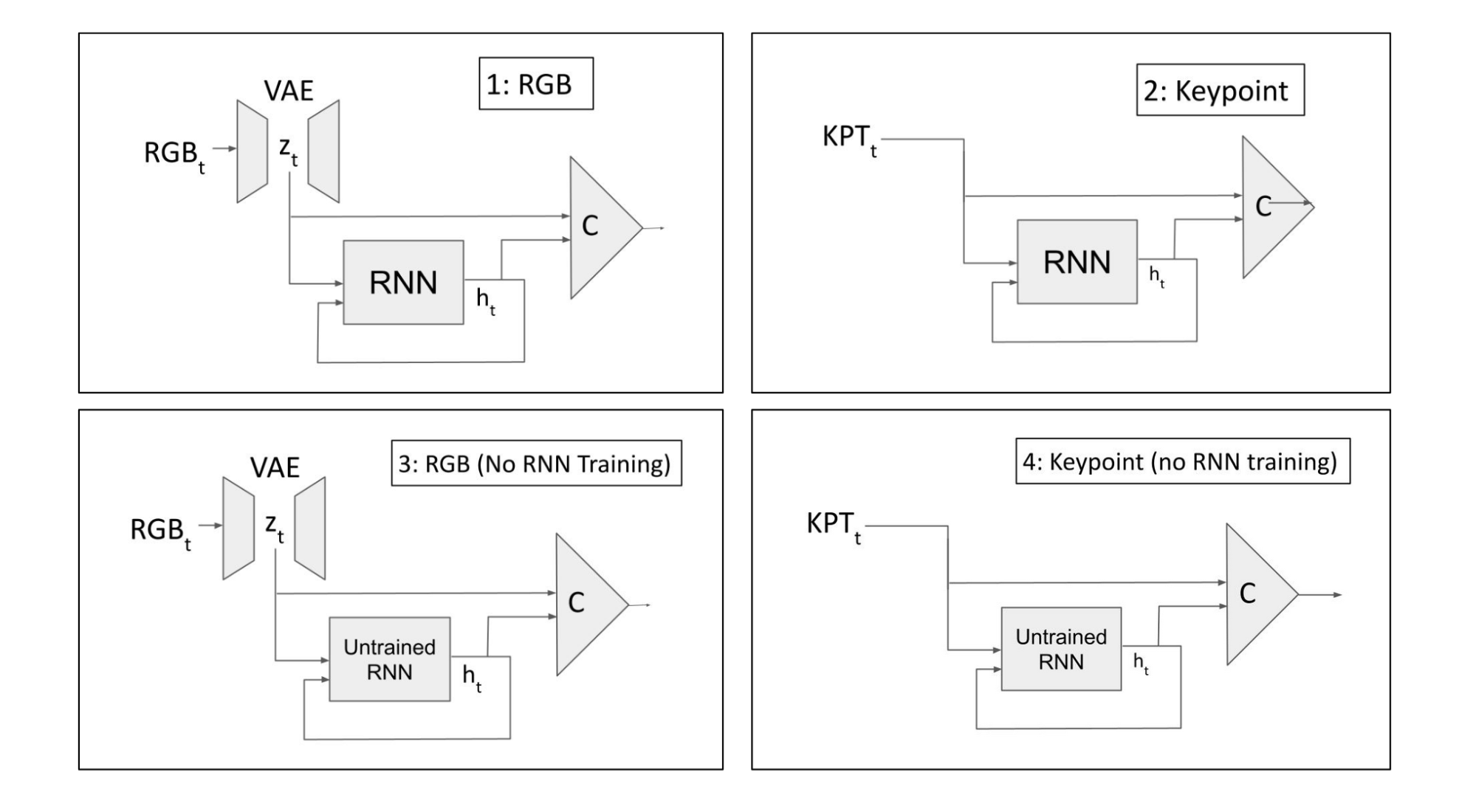}
\end{minipage}
\begin{minipage}{.9\textwidth}\centering
\includegraphics[width=.96\textwidth]{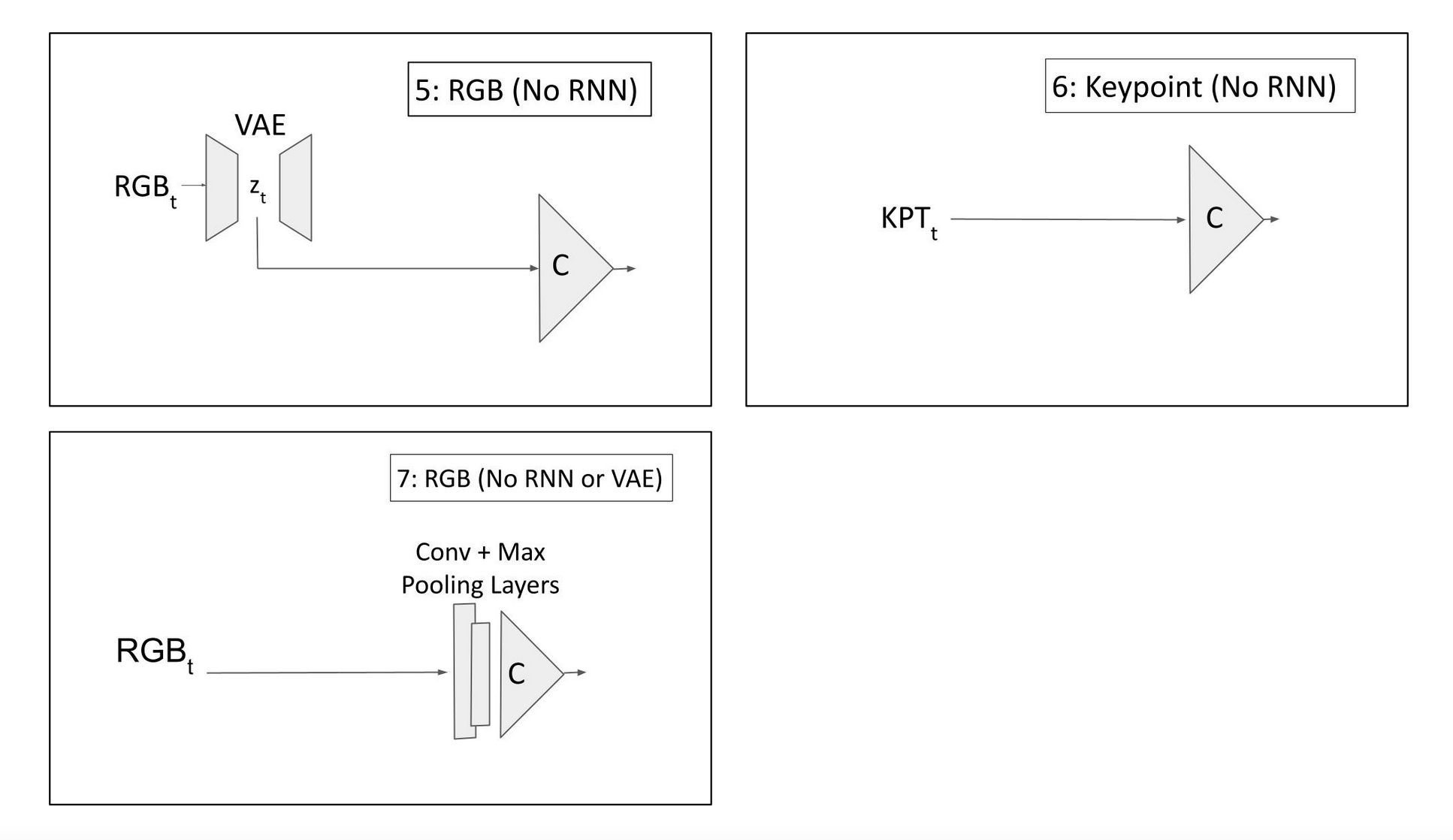}
\end{minipage}
\caption{Variations of and ablations on our world model approach.}
\label{fig:variations}
\end{figure}

We compare seven different variations of the world model approach for solving a cloth folding task in simulation (Figure \ref{fig:variations}). We evaluate the full pipeline using both keypoints and a VAE-learned reduced state representation of RGB images trained over 20 system training iterations. We compare this to a variation of the system in which an MDN-RNN is employed, but not trained, to evaluate whether the learned system dynamics improve the learned cloth-folding policy, or whether any benefit is solely due to the non-markovian information propagated through the hidden state even in an untrained RNN. 

To evaluate the impact on sample efficiency, we compare our model to an ablation removing the MDN-RNN from both models. In the case of the keypoint representation, the controller is trained directly on a vector concatenating the XYZ position of each keypoint and the XYZ position of the two robot grippers, for a length 78 vector representing the state. For the RGB representation, we compare two possible ablations - one where the controller is trained solely on the latent representation of the image state as produced by a VAE trained on the initial 1500 random rollouts in the environment, and one where two convolutional and max-pooling layers are added to the controller, bypassing the VAE to train the controller on the images directly. Because these models did not require storing and retraining on the rollouts collected at each controller training iteration, there were no memory limits associated with how long we were able to train them - we train each for 40 training loops in order to compare their sample efficiency to our model. 

For all controller training, we use a population size of 3 possible controllers, whose returns are averaged over 4 rollouts. The best controller is passed to the next generation.

\section*{Results and Discussion}

\begin{figure}[H]
    \centering
    \includegraphics[width=.8\textwidth]{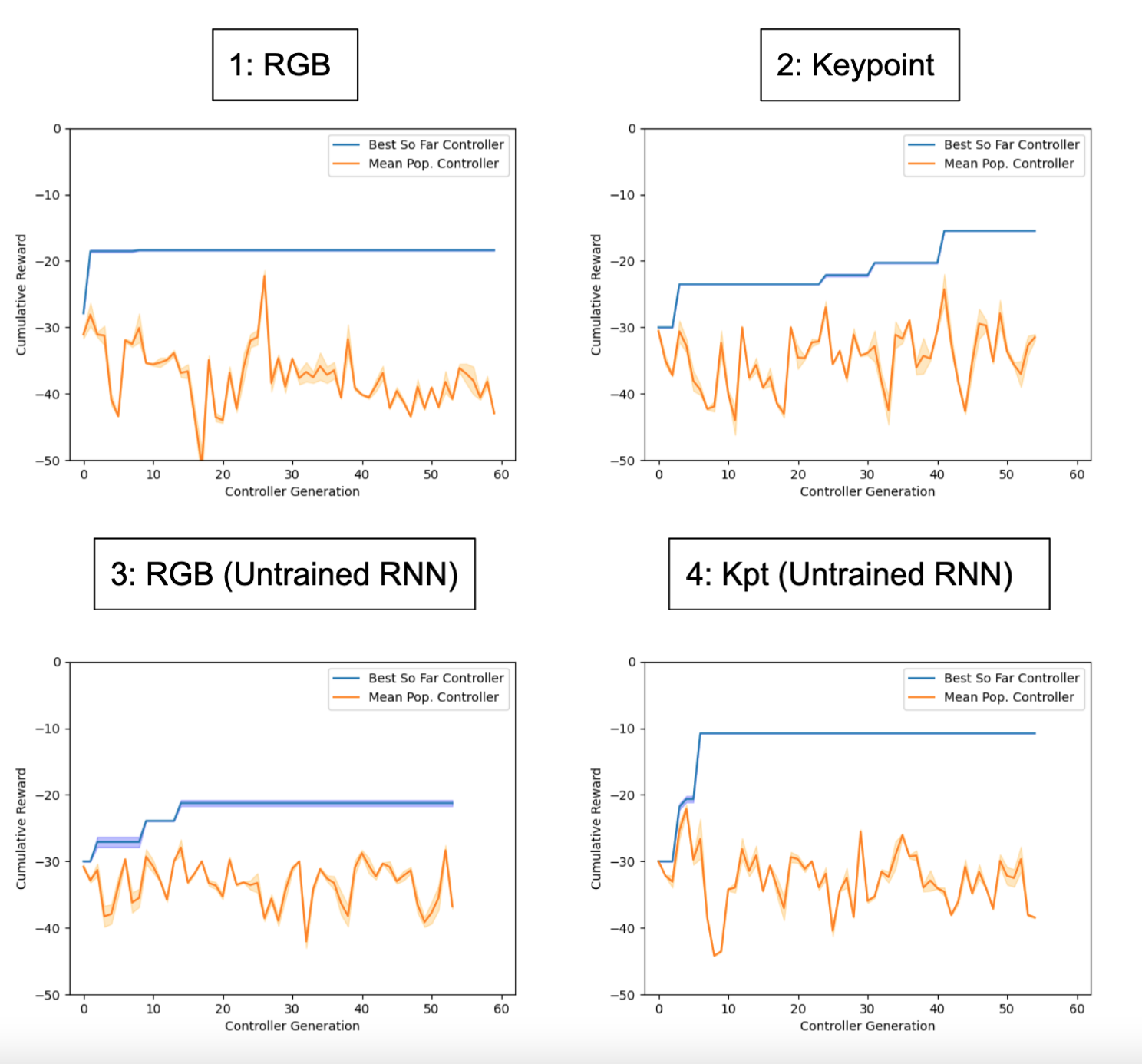}
    \caption{Results. The blue line represents the best controller achieved so far across all generations. The orange line represents the mean and 95\% confidence interval on the return from each subsequent generation.}
    \label{fig:results}
\end{figure}

We found that the keypoint model was consistently able to achieve a more optimal policy than the RGB model as illustrated in Figure \ref{fig:results}. We suspect that this is due to the component of the keypoint representation that includes the gripper positions - the VAE consistently struggles to represent the gripper positions accurately in the RGB state encoding, possibly due to the relatively small dimensionality of the latent space, and possibly exacerbated by the grippers being similar in color to the environment background. The VAE predictions are illustrated in Figure \ref{fig:appendix} in the appendix. Both models were able to achieve approximately the same controller performance on similar timescales with an untrained RNN (Figure \ref{fig:results} - 3, 4). This reflects the results in Ha and Schmidhuber \cite{Jurgen} on a continuous control racecar task, suggesting that for this task, the accuracy of the dynamics model is less important to performance than the simple presence of some form of non-markovian state information accessible to the controller. 

\begin{figure}[H]
 \begin{minipage}{.5\textwidth}
 \centering
\includegraphics[width=.9\linewidth]{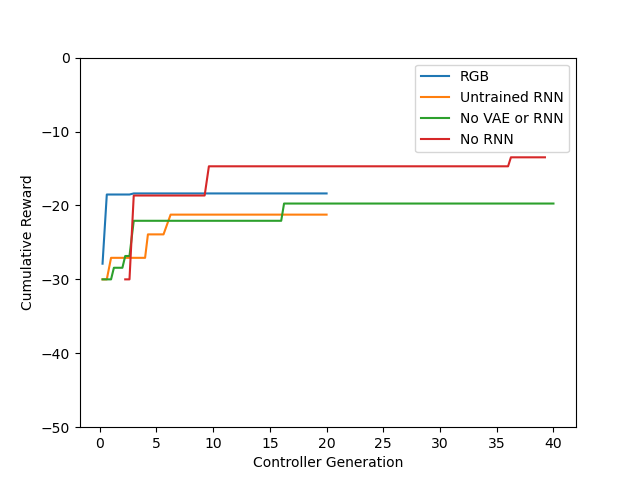}
\end{minipage}
 \begin{minipage}{.5\textwidth}
 \centering
\includegraphics[width=.9\linewidth]{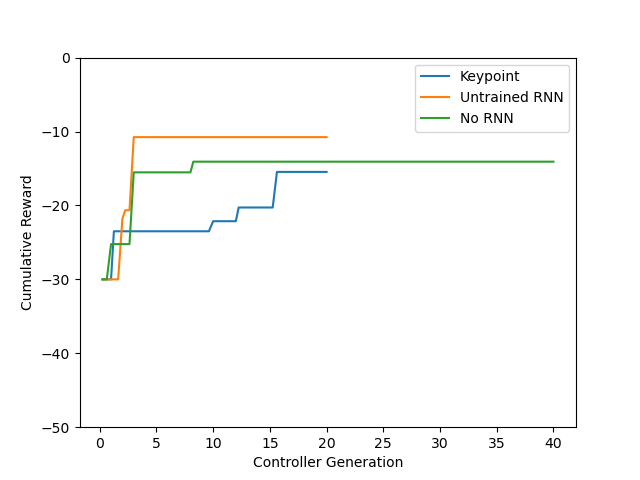}
\end{minipage}
\caption{Performance comparison of ablation experiments for rgb data (left). Performance comparison of ablation experiments for keypoints (right).}
\label{fig:sampleff}
\end{figure}

\begin{figure}[H]
\includegraphics[width=\textwidth]{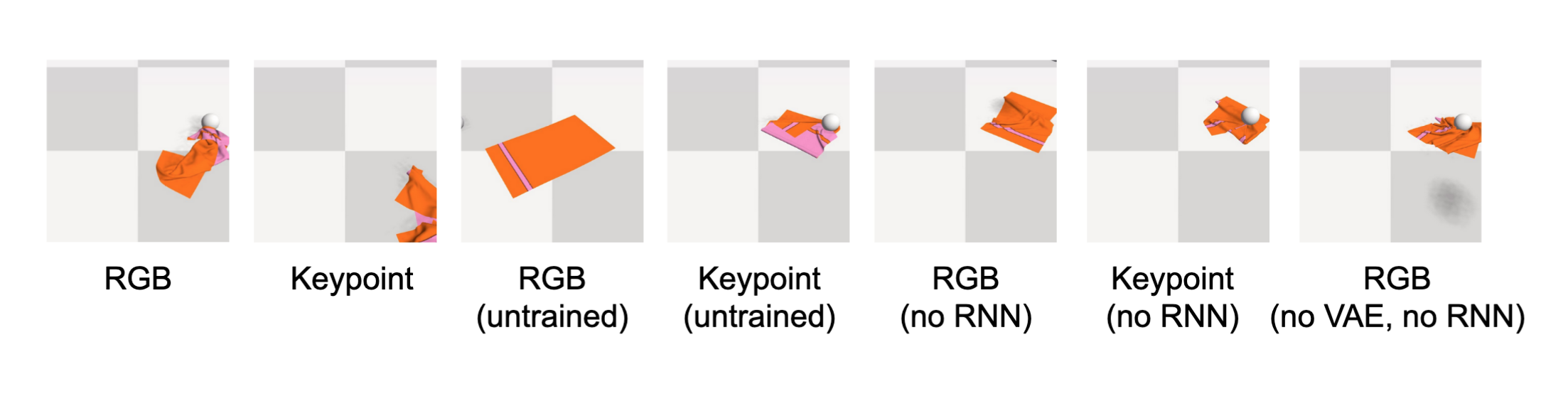}
\caption{The final state of the cloth after running the best controller output by each experiment. Untrained refers to the model with untrained RNN.}
\label{fig:controller}
\end{figure}

To evaluate the impact of the model on sample efficiency, we compare the best controller achieved over 40 generations between our model and our ablations in Figure \ref{fig:sampleff}. The controller executions are illustrated in Figure \ref{fig:controller}. We find that sample efficiency and performance are both improved by the use of a reduced dimensionality representation-whether the use of keypoints, or the learned VAE state encoding-but the presence or absence of the RNN did not provide any notable gains. 

One possible reason for this is failure of the MDN-RNN to accurately learn the dynamics of the environment. As shown in the MDN-RNN predictions figure (Figure \ref{fig:appendix}) in the appendix, the model fails to learn important information about cloth dynamics in both the keypoint and RGB domains. This did not improve substantially even when trained on more relevant data over the course of multiple training iterations, and may be impacted by the difficulty of the environment, or limitations of the representational power of the model we used. Further exploration into alternate approaches would be useful in establishing whether the use of a more effective RNN in learning a world model could improve performance on this task. 

\section*{Conclusion}
In this project, we investigated the use of world models and structured data in robotic cloth folding task, evaluating their effect on sample efficiency for robotics domain. We found that the learning state representation through VAE is advantageous but the use of MDN-RNN does not improve the performance for our task, given that our implementation is not current state-of-the-art, and our experimentation was limited by computing constraints. In addition using structured data, i. e. keypoints instead of RGB images, improved both performance and sample efficiency more, indicating that sample efficiency of world models can be increased with the use of a structured feature space. However, deployment of world models in real-world robotics tasks seems to require additional efforts. 

\bibliographystyle{plainnat}
\bibliography{references}

\begin{thebibliography}{16}
\providecommand{\natexlab}[1]{#1}
\providecommand{\url}[1]{\texttt{#1}}
\expandafter\ifx\csname urlstyle\endcsname\relax
  \providecommand{\doi}[1]{doi: #1}\else
  \providecommand{\doi}{doi: \begingroup \urlstyle{rm}\Url}\fi

\bibitem[Arriola-Rios et~al.(2020)Arriola-Rios, Guler, Ficuciello, Kragic,
  Siciliano, and Wyatt]{structure}
Veronica~E. Arriola-Rios, Puren Guler, Fanny Ficuciello, Danica Kragic, Bruno
  Siciliano, and Jeremy~L. Wyatt.
\newblock Modeling of deformable objects for robotic manipulation: A tutorial
  and review.
\newblock \emph{Frontiers in Robotics and AI}, 7:\penalty0 82, 2020.
\newblock ISSN 2296-9144.
\newblock \doi{10.3389/frobt.2020.00082}.
\newblock URL
  \url{https://www.frontiersin.org/article/10.3389/frobt.2020.00082}.

\bibitem[de~Boer et~al.(2004)de~Boer, Kroese, Mannor, and
  Rubinstein]{Boer04atutorial}
Pieter-Tjerk de~Boer, Dirk~P. Kroese, Shie Mannor, and Reuven~Y. Rubinstein.
\newblock A tutorial on the cross-entropy method.
\newblock \emph{ANNALS OF OPERATIONS RESEARCH}, 134, 2004.

\bibitem[Dulac-Arnold et~al.(2019)Dulac-Arnold, Mankowitz, and
  Hester]{dulacarnold2019challenges}
Gabriel Dulac-Arnold, Daniel Mankowitz, and Todd Hester.
\newblock Challenges of real-world reinforcement learning, 2019.

\bibitem[Florence et~al.(2018)Florence, Manuelli, and
  Tedrake]{florence2018dense}
Peter~R. Florence, Lucas Manuelli, and Russ Tedrake.
\newblock Dense object nets: Learning dense visual object descriptors by and
  for robotic manipulation, 2018.

\bibitem[Gao and Tedrake(2021)]{gao2021kpam}
Wei Gao and Russ Tedrake.
\newblock kpam 2.0: Feedback control for category-level robotic manipulation,
  2021.

\bibitem[Ha and Schmidhuber(2018)]{Jurgen}
David Ha and J{\"{u}}rgen Schmidhuber.
\newblock World models.
\newblock \emph{CoRR}, abs/1803.10122, 2018.
\newblock URL \url{http://arxiv.org/abs/1803.10122}.

\bibitem[Hafner et~al.(2020{\natexlab{a}})Hafner, Lillicrap, Ba, and
  Norouzi]{hafner2020dream}
Danijar Hafner, Timothy Lillicrap, Jimmy Ba, and Mohammad Norouzi.
\newblock Dream to control: Learning behaviors by latent imagination,
  2020{\natexlab{a}}.

\bibitem[Hafner et~al.(2020{\natexlab{b}})Hafner, Lillicrap, Norouzi, and
  Ba]{Dreamerv2}
Danijar Hafner, Timothy~P. Lillicrap, Mohammad Norouzi, and Jimmy Ba.
\newblock Mastering atari with discrete world models.
\newblock \emph{CoRR}, abs/2010.02193, 2020{\natexlab{b}}.
\newblock URL \url{https://arxiv.org/abs/2010.02193}.

\bibitem[Hansen(2016)]{hansen2016cma}
Nikolaus Hansen.
\newblock The cma evolution strategy: A tutorial, 2016.

\bibitem[Kingma and Welling(2019)]{Diederik2019}
Diederik~P. Kingma and Max Welling.
\newblock An introduction to variational autoencoders.
\newblock \emph{CoRR}, abs/1906.02691, 2019.
\newblock URL \url{http://arxiv.org/abs/1906.02691}.

\bibitem[Kipf et~al.(2020)Kipf, van~der Pol, and Welling]{kipf2020contrastive}
Thomas Kipf, Elise van~der Pol, and Max Welling.
\newblock Contrastive learning of structured world models, 2020.

\bibitem[Kulkarni et~al.(2019)Kulkarni, Gupta, Ionescu, Borgeaud, Reynolds,
  Zisserman, and Mnih]{kulkarni2019unsupervised}
Tejas Kulkarni, Ankush Gupta, Catalin Ionescu, Sebastian Borgeaud, Malcolm
  Reynolds, Andrew Zisserman, and Volodymyr Mnih.
\newblock Unsupervised learning of object keypoints for perception and control,
  2019.

\bibitem[Lin et~al.(2020)Lin, Wang, Olkin, and Held]{softgym}
Xingyu Lin, Yufei Wang, Jake Olkin, and David Held.
\newblock Softgym: Benchmarking deep reinforcement learning for deformable
  object manipulation.
\newblock \emph{CoRR}, abs/2011.07215, 2020.
\newblock URL \url{https://arxiv.org/abs/2011.07215}.

\bibitem[Sekar et~al.(2020)Sekar, Rybkin, Daniilidis, Abbeel, Hafner, and
  Pathak]{sekar2020planning}
Ramanan Sekar, Oleh Rybkin, Kostas Daniilidis, Pieter Abbeel, Danijar Hafner,
  and Deepak Pathak.
\newblock Planning to explore via self-supervised world models, 2020.

\bibitem[Sutton and Barto(2018)]{sutton2018reinforcement}
Richard~S Sutton and Andrew~G Barto.
\newblock \emph{Reinforcement learning: An introduction}.
\newblock MIT press, 2018.

\bibitem[Tassa et~al.(2018)Tassa, Doron, Muldal, Erez, Li, de~Las~Casas,
  Budden, Abdolmaleki, Merel, Lefrancq, Lillicrap, and
  Riedmiller]{tassa2018deepmind}
Yuval Tassa, Yotam Doron, Alistair Muldal, Tom Erez, Yazhe Li, Diego
  de~Las~Casas, David Budden, Abbas Abdolmaleki, Josh Merel, Andrew Lefrancq,
  Timothy Lillicrap, and Martin Riedmiller.
\newblock Deepmind control suite, 2018.

\end{thebibliography}

\section*{Appendix}
\label{sec:App}

\begin{figure}[H]
\includegraphics[width=.97\textwidth]{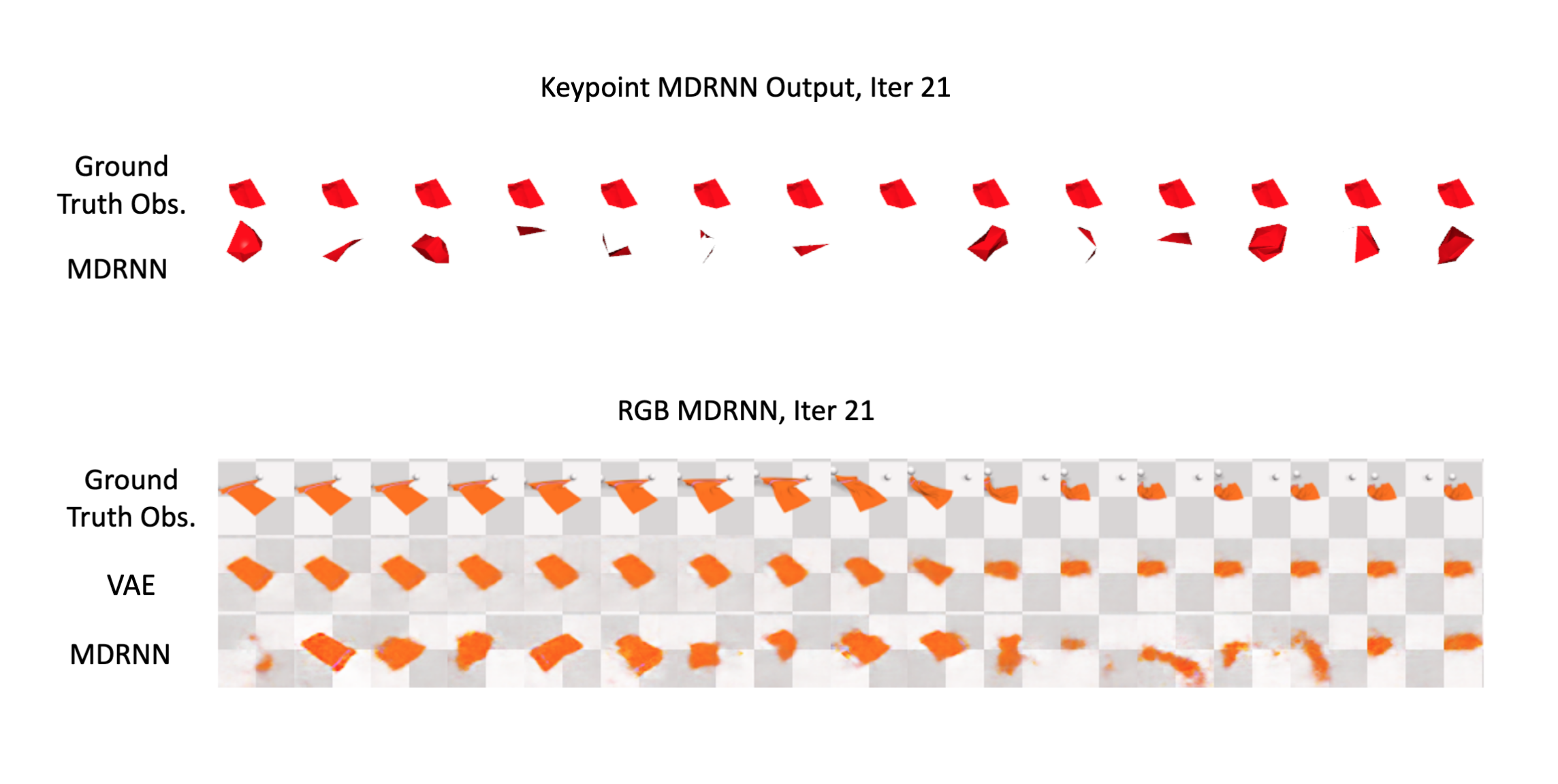}
\caption{The output of the RGB and Keypoint MDN-RNNs after training. The top row shows the ground truth (in the keypoint case, a mesh constructed from the keypoint positions), the middle row for RGB shows the VAE reconstruction for that state, and the bottom row shows the reconstruction of the MDN-RNN prediction for that state, given the previous state. In both cases, the MDN-RNN seems to struggle to capture accurate dynamics in the environment.}
\label{fig:appendix}
\end{figure}

\end{document}